\pdfoutput=1

\documentclass[11pt]{article}

\usepackage[final]{acl}
\usepackage{enumitem}
\setlist[itemize]{leftmargin=*}
\usepackage{tabularx}
\usepackage{graphicx}
\usepackage{times}
\usepackage{latexsym}
\usepackage{booktabs}
\usepackage{amsmath}
\usepackage{amsfonts}

\usepackage[T1]{fontenc}

\usepackage[utf8]{inputenc}
\usepackage[T1]{fontenc}
\usepackage{CJKutf8}

\usepackage{microtype}

\usepackage{inconsolata}

\usepackage{listings}
\usepackage{subcaption}
\usepackage{tablefootnote}
\usepackage[most]{tcolorbox}
\usepackage{fancyhdr}
\usepackage{lipsum}

\tcbuselibrary{breakable}

\title{Efficient Medical VIE via Reinforcement Learning}

\author{
  \textbf{Lijun Liu\textsuperscript{1,}\textsuperscript{*}},
  \textbf{Ruiyang Li\textsuperscript{1,}\textsuperscript{*}},
  \textbf{Zhaocheng Liu\textsuperscript{1,}\textsuperscript{\textdagger}}
  \\
  \textbf{Chenglin Zhu\textsuperscript{1,2}},
  \textbf{Chong Li\textsuperscript{1,2}},
  \textbf{Jiehan Cheng\textsuperscript{1,3}},
  \textbf{Qiang Ju\textsuperscript{1}},
  \textbf{Jian Xie\textsuperscript{1}}
  \\
  \textsuperscript{1}Baichuan Inc. \ \ 
  \textsuperscript{2}Peking University \ \ 
  \textsuperscript{3}Renmin University of China
  \\
  \small{
    \textbf{Correspondence:} \href{mailto:lio.h.zen@gmail.com}{lio.h.zen@gmail.com}
  }
}
\begin{document}
\begin{CJK}{UTF8}{gbsn}
\maketitle
\begin{abstract}
Visual Information Extraction (VIE) converts unstructured document images into structured formats like JSON, critical for medical applications such as report analysis and online consultations.
Traditional methods rely on OCR and language models, while end-to-end multimodal models offer direct JSON generation.
However, domain-specific schemas and high annotation costs limit their effectiveness in medical VIE.
We base our approach on the Reinforcement Learning with Verifiable Rewards (RLVR) framework to address these challenges using only 100 annotated samples.
Our approach ensures dataset diversity, a balanced precision-recall reward mechanism to reduce hallucinations and improve field coverage, and innovative sampling strategies to enhance reasoning capabilities.
Fine-tuning Qwen2.5-VL-7B with our RLVR method, we achieve state-of-the-art performance on medical VIE tasks, significantly improving F1, precision, and recall.
While our models excel on tasks similar to medical datasets, performance drops on dissimilar tasks, highlighting the need for domain-specific optimization.
Case studies further demonstrate the value of reasoning during training and inference for VIE.
\end{abstract}

\section{Introduction}

Visual Information Extraction (VIE) \cite{wan2024omniparser,kuang2023visual,hong2022bros,kim2022ocr} aims to generate structured information, such as JSON, from unstructured document images. This capability is crucial for various medical applications such as report interpretation \cite{li2024improving} and online consultations \cite{liu2025exploring}.
The most common approach involves first applying Optical Character Recognition (OCR) \cite{feng2025dolphin,poznanski2025olmocr,wei2024general} to extract text, followed by leveraging large language models (LLMs) to extract and organize the text into a JSON structure. Additionally, end-to-end methods \cite{wan2024omniparser,bai2025qwen2,kuang2023visual,kim2022ocr} have emerged, including multimodal large models that directly output JSON from image inputs.

However, VIE tasks are highly domain-specific, with each domain requiring customized schemas. \cite{park2019cord,huang2019icdar2019} The keys and values within these schemas are often defined by intricate domain-specific details, posing significant challenges for applying general-purpose VIE models to specialized fields. This aspect fundamentally differentiates structured VIE from OCR. Moreover, the annotation cost for VIE tasks is relatively high. These challenges have resulted in suboptimal performance of existing methods in medical VIE scenarios.

Given the nontrivial relationship between diverse image inputs and outputs conforming to predefined schemas, we argue that VIE models need reasoning capabilities \cite{openaio1} to address these complexities.
To mitigate the high annotation cost, we explore efficient training paradigms using only 100 annotated samples.
Combining these two considerations, we adopt Reinforcement Learning with Verifiable Rewards (RLVR) \cite{guo2025deepseek,team2025kimi} to achieve efficient medical visual extraction.

Specifically, our design within the RLVR framework focuses on three key aspects.
First, we ensure diversity in the 100 image samples to make the dataset representative and varied.
Second, we carefully design the reward mechanism by incorporating a weighted combination of precision and recall, where precision reduces model hallucinations and recall ensures the model captures all the predefined fields of interest.
Lastly, we adopt two sampling strategies: one requires each response to include all fields, with rewards calculated against the ground truth for all fields, while the other evaluates responses using a random subset of fields from the total schema.
By integrating these carefully designed components, we aim to establish an efficient and robust solution for medical VIE tasks.

Based on our proposed method, we finetuned Qwen2.5-VL-7B to have VIE RLVR models. 
We evaluate the models on medical and general VIE tasks. 
Our VIE RLVR models achieve SOTA performance on F1, precision and recall metrics on medical VIE tasks, indicating the advantage of our proposed method. 
We chose four widely-considered general VIE tasks for further evaluation.
On two tasks that are similar to the medical report dataset, our models highly outperforms Qwen2.5-VL-7B.
Meanwhile on other two tasks that are much different with our medical report dataset, our models failed to outperforms the base model, revealing the significant gap between different VIE tasks. 
The comparison of VIE models with and without thinking process while training and inferring are also delivered. In our case studies one can see how model benefits from thinking in dealing with VIE tasks.

\section{Related Work}

\subsection{Visual Information Extraction}
Visual Information Extraction (VIE) converts unstructured document images into structured outputs (e.g., key–value pairs or JSON), supporting applications like receipt understanding, form parsing, and medical document analysis\cite{huang2022layoutlmv3,powalski2021going,appalaraju2021docformer}. Existing methods fall into two main types: two-stage approaches that apply OCR followed by language models for structural parsing\cite{xu2020layoutlm,xu2020layoutlmv2}, and end-to-end models that directly generate outputs from images without OCR \cite{kim2022ocr,zhang2020trie}. Though effective on low-complexity benchmarks such as FUNSD, SROIE, and CORD \cite{Jaume2019FUNSD,Huang2019SROIE,park2019cord,cao2022query,wang2021towards}, these models often omit required fields, hallucinate content, and generalize poorly to unseen layouts—especially under few-shot or domain-shift conditions. These issues are exacerbated in the medical domain\cite{ma2023extracting,zheng2022improving}, where layouts vary widely and annotated data is scarce. While recent advances like layout-aware pretraining \cite{chen2022xdoc,adnan2024layoutlmv3,luo2023geolayoutlm}, graph-based models \cite{yu2021pick}, and schema-guided prompting \cite{wangchatschema,li2024improving,yao2024learning} provide partial solutions, they often fall short of ensuring both structural completeness and semantic accuracy under low-resource constraints.
\subsection{Reinforcement Learning for MLLM Reasoning}
Reinforcement Learning (RL) has emerged as a pivotal research direction for enhancing the complex reasoning capabilities of LLMs \cite{guo2025deepseek,jaech2024openai,shao2024deepseekmath,hui2024qwen2, ying2024internlm}. OpenAI-o1 \cite{jaech2024openai} adopted Reinforcement Learning from Human Feedback (RLHF) during the fine-tuning process, significantly enhancing the model's reasoning abilities and its alignment with human preferences. More recently, DeepSeek-R1 \cite{guo2025deepseek} employed GRPO \cite{shao2024deepseekmath}, which, unlike traditional RL algorithms dependent on critic models, directly utilizes rule-based verifiable rewards to guide the model's reasoning process. This approach has greatly simplified the training procedure and proven highly effective in improving reasoning capabilities. This trend is gradually extending to MLLMs to further enhance their visual reasoning abilities \cite{xu2024llava,liu2025seg,yu2025perception,yang2025r1,zhou2503r1}. Studies such as Visual-RFT \cite{liu2025visual} and VLM-R1 \cite{shen2025vlm} have shown that for single-image visual grounding tasks, direct application of few-shot GRPO can achieve improvements surpassing supervised fine-tuning. The GoT-R1 \cite{duan2025got} framework applies RL to enhance semantic spatial reasoning in visual generation. Vision-R1 \cite{huang2025vision} enhances multimodal mathematical reasoning capabilities by using DeepSeek-R1 to augment multimodal Chain of Thought (CoT) datasets and adopting step-by-step thought inhibition during GRPO training. In this research, we aim to extend this paradigm to the Medical VIE tasks mentioned earlier.
\section{Method} 
\begin{figure}[t]
  \includegraphics[width=\columnwidth]{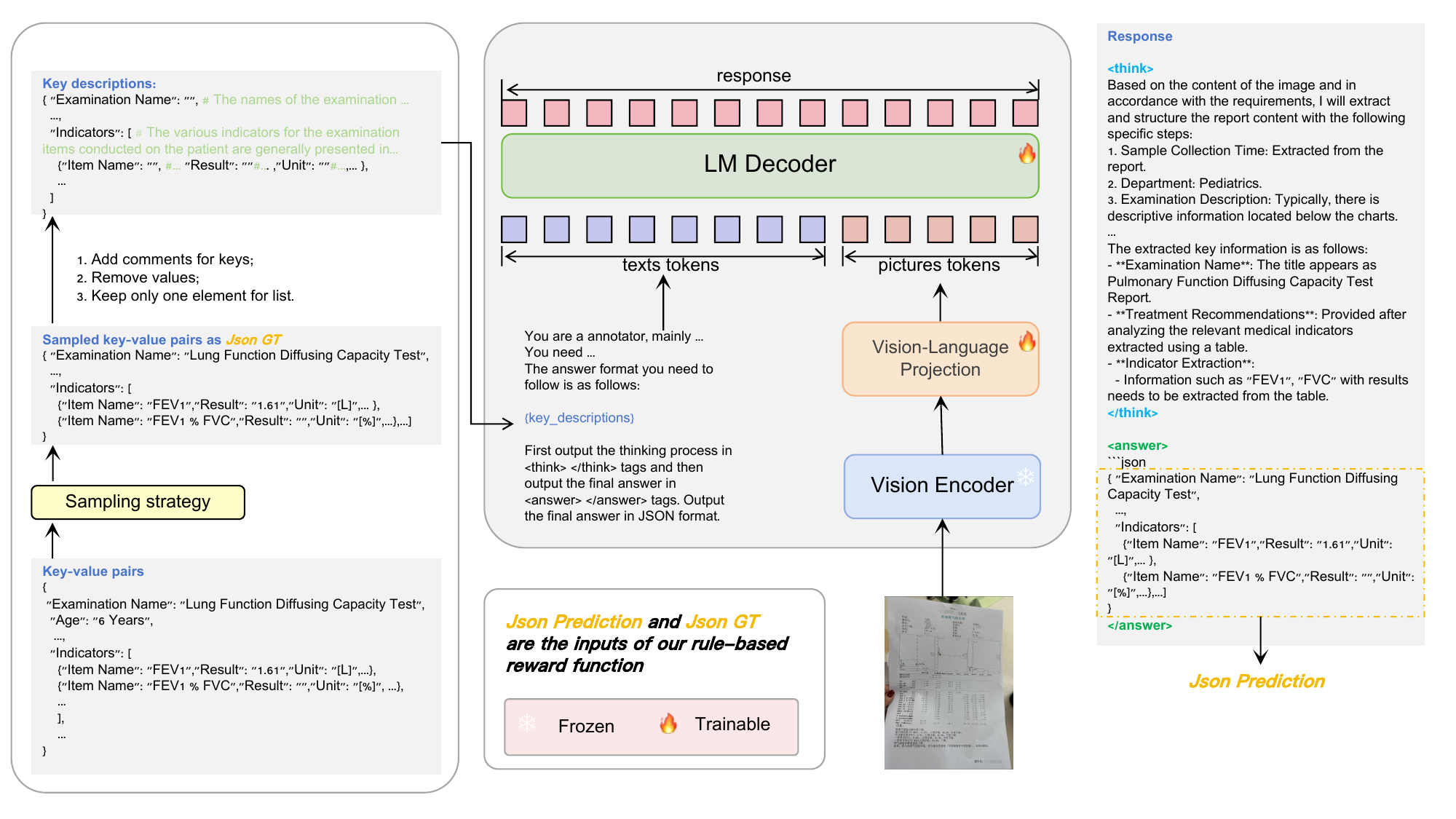}
  \caption{\textbf{Overview of our proposed method}.
    A sampling strategy is employed to randomly select key-value pairs from the annotated JSON data associated with each image, serving as the ground truth. The descriptions of these keys are incorporated into a predefined prompt to generate queries. These queries, along with the corresponding images, are input into the model to produce responses. The JSON predictions are then extracted from the responses, and rewards are computed by comparing them with the ground truth.}
  \label{fig:experiments}
\end{figure}

\subsection{Preliminary}
Due to the intricate relationship between heterogeneous image inputs and outputs that adhere to predefined schemas, we argue that VIE models must possess reasoning abilities to effectively manage these complexities.
In contrast to approaches that explicitly replicate intermediate reasoning steps, RLVR \cite{guo2025deepseek,team2025kimi} relies solely on outcome-driven feedback, facilitating scalable reinforcement learning across extensive task datasets.

Group Relative Policy Optimization (GRPO) \cite{guo2025deepseek} is an efficient RL algorithm that eliminates the need for a separate critic model.
Given a query $q$, GRPO samples a group of $G$ outputs $\{o_1, o_2, \dots, o_G\}$ from the old policy $\pi_{\theta_{\text{old}}}$.
These outputs are evaluated using reward functions to obtain individual rewards $\{r_1, r_2, \dots, r_G\}$.
The advantage is computed by normalizing the rewards within the group:
\begin{equation}
    A_i = \frac{r_i - \mathrm{mean}(\{r_1, r_2, \dots, r_G\})}{\mathrm{std}(\{r_1, r_2, \dots, r_G\})}. \\
\end{equation}
Then the policy is updated by optimizing the following objective:
\begin{align}
  \label{eq:grpo_obj_widetext}
  &\mathcal{J}_{\text{GRPO}}(\theta) = \mathbb{E}_{q \sim \mathcal{D}, \{o_i\}_{i=1}^G \sim \pi_{\theta_{\text{old}}}} \notag\\
  &\frac{1}{G} \sum_{i=1}^{G} \frac{1}{|o_i|} \sum_{t=1}^{|o_i|} 
  \Bigg( 
\min \bigg( 
\varphi_{i,t}(\theta) A_{i,t},\ 
\text{clip} (\varphi_{i,t}(\theta),\ 1-\epsilon,\ 1+\epsilon)A_{i,t} \bigg) - \beta \, \mathbb{D}_{\text{KL}} \left[ \pi_\theta \| \pi_{\text{ref}} \right]
\Bigg),
\end{align}
where
\begin{equation}
    \varphi_{i,t}(\theta) = \frac{\pi_\theta(o_{i,t}\ |\ q,o_{i,<t})}{\pi_{\theta_\text{old}}(o_{i,t}\ |\ q,o_{i,<t})}. \\
\end{equation}

Additionally, we adopt several key techniques from DAPO \cite{yu2025dapo}, including Clip-Higher and Token-Level Policy Gradient Loss.
With the introduction of the two, the objective function undergoes some slight modifications as follows:
\begin{align}
  \label{eq:grpo_obj_widetext}
  &\mathcal{J}_{\text{GRPO}}(\theta) = \mathbb{E}_{q \sim \mathcal{D}, \{o_i\}_{i=1}^G \sim \pi_{\theta_{\text{old}}}} \notag\\
  &\color{red}{\frac{1}{\sum_{i=1}^G|o_i|} \sum_{i=1}^{G} \sum_{t=1}^{|o_i|}}\color{black}
  \Bigg( 
\min \bigg( 
\varphi_{i,t}(\theta) A_{i,t},\ 
\text{clip} (\varphi_{i,t}(\theta),\ 1-\epsilon_{\color{red}{\text{low}}},\ 1+\epsilon_{\color{red}{\text{high}}})A_{i,t} \bigg) - \beta \, \mathbb{D}_{\text{KL}} \left[ \pi_\theta \| \pi_{\text{ref}} \right]
\Bigg).
\end{align}

\subsection{Image Diversity}
We collected over 17,000 medical domain images along with their corresponding OCR ground truth.
These images encompass a diverse range of report types, including laboratory reports (e.g., blood, urine, and stool tests), diagnostic reports (e.g., endoscopy, electrocardiograms, ultrasounds, and CT scans), and pathological reports (e.g., biopsy analyses and tumor staging).
Furthermore, the diversity of the images extends to factors such as shooting angles, creases in the reports, the presence of obstructions, handwritten elements (e.g., doctor signatures), and varying backgrounds in the photographs.

From this dataset, we manually selected 100 images that exhibit high diversity across these dimensions.
Using GPT-4o \cite{openai4o}, the OCR ground truth was converted into JSON format based on a predefined medical schema (see Appendix \ref{sec:appendix_pms}).
The JSON outputs were then manually reviewed and corrected to produce the final JSON ground truth.

\subsection{Rule-based Reward Mechanism}
We design a rule-based reward function to optimize the model's ability to generate JSON outputs by measuring similarity with the ground truth.
The reward computation consists of the following steps: 

\textbf{Format Score}. We generally adopt the format of R1-Zero \cite{guo2025deepseek}, which includes two components: \colorbox{lightgray}{think} and \colorbox{lightgray}{answer}. The format score $r_\text{format}$ is 1 if both components meet the required specifications; otherwise, $r_\text{format}$ is 0. 

\textbf{JSON Preprocessing}. Parse the JSON object from the answer and flatten it into a non-nested key-value dictionary.
Specifically, this involves traversing all leaf nodes in the JSON structure.
Each leaf node's key in the dictionary is formed by concatenating the keys along the path from the root to the leaf, and its corresponding value is the value of the leaf node.
Given the model output $\hat{y}$ and the ground truth $y$, the preprocessing step converts them into $S_p$ and $S_g$, respectively.

\begin{figure}[t]
  \includegraphics[width=\columnwidth]{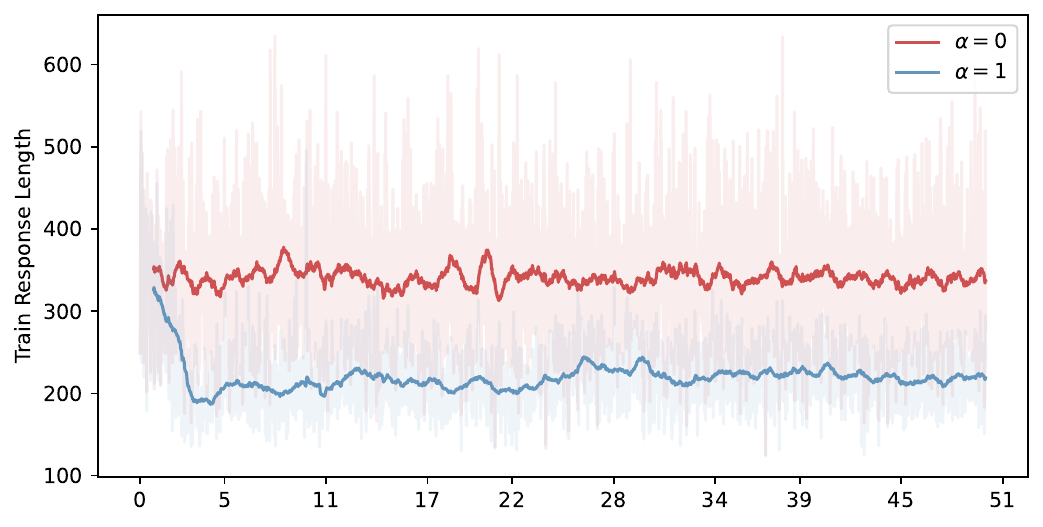}
  \caption{\textbf{Impact of the hyperparameter $\alpha$ on response length when the Sampling Strategy is enabled}.
    The semi-transparent and the solid lines indicate raw samples and the smoothed trend.}
  \label{fig:alpha}
\end{figure}
\textbf{Matching Score}. The similarity between $S_p$ and $S_g$ is measured through a weighted combination of precision and recall.
We define $n_{\text{matched}}$ as the number of correctly matched key-value pairs between $S_p$ and $S_g$.
Accordingly, precision and recall are defined as $\frac{n_\text{matched}}{|S_p|}$ and $\frac{n_\text{matched}}{|S_g|}$, respectively.
Therefore, the matching score is defined as:
\begin{equation}\label{eq:matching}
    r_\text{matching} = 
\begin{cases} 
\alpha \frac{n_\text{matched}}{|S_p|} + (1 - \alpha) \frac{n_\text{matched}}{|S_g|} & \text{if } |S_p| > 0, \\
0 & \text{else}.
\end{cases}
\end{equation}
As shown in Figure \ref{fig:alpha}, $\alpha$ serves as a critical hyperparameter to balance precision and recall during optimization:
\begin{itemize}
    \item When $\alpha$ equals 1, the reward function focuses solely on precision, allowing the model to achieve 100\% precision by outputting just a single perfectly matched key-value pair.
    \item when $\alpha$ equals 0, the reward function emphasizes recall, potentially causing the model to generate numerous hallucinated key-value pairs in an attempt to retrieve all key-value pairs from $S_g$.
\end{itemize}

\textbf{Final Reward Score}. The reward for the i-th sample is calculated as the sum of the format score and the matching score, expressed as:
\begin{equation}
    r_i = r_\text{format} + r_\text{matching}
\end{equation}

\textbf{Comparison with SFT}.
Compared to SFT, our proposed reward function better accommodates the unordered nature of JSON data.
Specifically, the unordered property of JSON allows a single image to correspond to multiple ground truths.
SFT uses cross-entropy loss on fixed JSON ground truths during training, which may lead to data ambiguity and affect model performance.

\subsection{Sampling Strategy}
\begin{figure}[t]
  \includegraphics[width=\columnwidth]{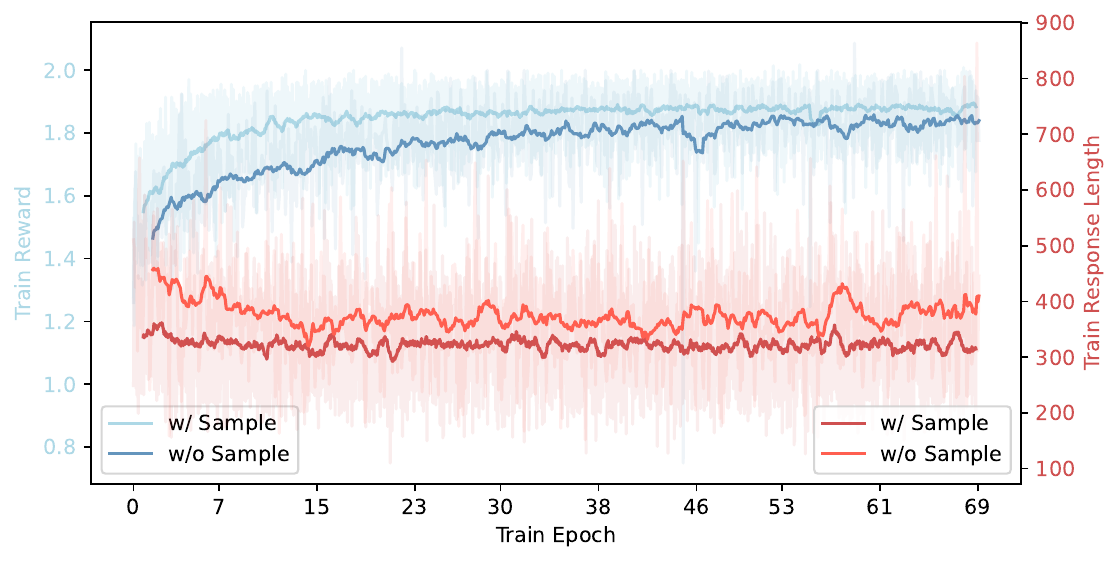}
  \caption{\textbf{Comparison of Reward and Response Length Trends Between Sampling and Non-Sampling Strategies During Training}. The semi-transparent and the solid lines indicate raw samples and the smoothed trend.}
  \label{fig:sample strategy}
\end{figure}
To assess the impact of query diversity on experimental outcomes, we employ two data construction strategies.
The first strategy involves random sampling of keys corresponding to the JSON data of an image, thereby generating varied queries.
The second strategy forgoes sampling, utilizing all keys, which results in all samples sharing a same query.
Observations from Figure \ref{fig:sample strategy} indicate that key sampling leads to shorter response lengths, as the number of keys decreases post-sampling, consequently shortening response lengths.
Additionally, from the reward curve, it can be observed that key sampling results in faster reward growth, which we attribute to the reduced number of keys making the training task simpler and thus accelerating reward acquisition.

\section{Experiments}

\subsection{VIE Metrics}

Various metrics are used for evaluation, including 
 the field-level precision, recall, F1 scores and tree-edit-distance(TED) based accuracy as in \cite{kim2021donut}. It is noted that TED based accuracy mainly reflects the correctness of trees' topology. In practical VIE scenarios, we pay more attention to indicators such as F1 score, precision, and recall, which also reflect the correctness of extracted text information.  
 \begin{itemize}
 \item TED based accuracy measures the degree of match between the model's output and the ground truth by calculating the edit distance between two tree structures, and the tree edit distance is used to quantify discrepancies between the predicted and actual structures.
 \item Field-level precision is the proportion of correctly extracted fields among all predicted fields, defined as  $n_\text{matched}/|S_p|$ in Eq. \ref{eq:matching}.
 \item Field-level recall is the proportion of correctly extracted fields among all actual fields, defined as $n_\text{matched}/|S_g|$ in Eq. \ref{eq:matching}.
 \item Field-level F1 score is the harmonic mean balancing precision and recall, measuring overall accuracy of field-specific extraction, respectively.
 \end{itemize}

\subsection{VIE Baselines}

To compare our results with existing works, we introduce models with various types and different outputs. 

\textbf{Pipeline models.} The pipeline models for OCR task are always composed of layout recognizer and OCR models for plain text, math functions and tables, and the OCR results are collected and rearranged into markdown, HTML or LaTeX format.
\begin{itemize}
\item MinerU \cite{wang2024mineru} is a widely used pipeline model for OCR tasks, it uses LayoutLMv3 \cite{huang2022layoutlmv3} or DocLayout-YOLO \cite{zhao2024doclayoutyoloenhancingdocumentlayout} for document layout detection, an YOLO-v8 model \footnote{https://yolov8.com/} for math function detection, UniMERNet \cite{wang2024unimernetuniversalnetworkrealworld} for math function recognition, RapidTable \footnote{https://github.com/RapidAI/RapidTable}, TableMaster \cite{paddleocr2020} or StructEqTable \cite{xia2024docgenomeopenlargescalescientific} for table recognition, PaddleOCR \cite{paddleocr2020} for plain text OCR and LayOutReader \cite{Pang_Faster_LayoutReader_based_2024} for reading order analysis. When evaluating MinerU, we align the version and parameter settings with those in OmniDocBench \cite{ouyang2025omnidocbench}.  
\item Marker \cite{wang2021pgnet} integrates several open source models to parse document, and we align the version and parameter settings of Marker as in OmniDocBench.
\end{itemize}

\textbf{Expert models.} GOT-OCR \cite{wei2024general} is a large multimodal model trained for document parsing, which firstly used a multi-stage training strategy to train an end-to-end OCR model.

\textbf{General MLLMs.} We include general purpose MLLMs such as GPT4o\cite{openai4o}, Qwen2.5-VL-7B\cite{bai2025qwen2}, Qwen2.5-VL-72B and InternVL-2.5-78B \cite{chen2025expandingperformanceboundariesopensource} as baselines. The usage and parameter settings of these models are aligned with those in OmniDocBench.

\subsection{Our VIE Models}

In this subsection we first introduce our VIE models finetuned by SFT and RLVR methods, and we report the implementation details of training.

\textbf{VIE SFT models.} We show the result of models trained by VIE SFT and compare them with the VIE RLVR finetuned models, in order to analyze the benefits given by RLVR instead of SFT in VIE tasks. 
\begin{itemize}
    \item JSON-SFT-100 model is finetuned with 100 high quality medical VIE data, model leans to extract key information by directly supervised finetuning.
    \item OCR-SFT-17K model is finetuned with 17K medical report OCR data. The model parse images into markdown format, and we apply GPT4o to rearrange the markdown into JSON while evaluating its VIE performance.
\end{itemize}

\textbf{VIE RLVR models.} Three VIE RLVR models are trained to evaluate our proposed method:   
\begin{itemize}
\item RL-100 is trained on 100 high quality data, during training the model extract information of randomly sampled fields. Note that the images of the 100 training dataset is same to JSON-SFT-100. 
\item RL-100(w/o sample) is trained with similar schedule and 100 data, but during training the model is required to extract all the key information in the images. 
\item OCR-SFT-17K-RL-100 model has the same RL stage with RL-100, and it is additionally supervised finetuned with 17K high quality OCR data composed of in-service medical reports and their manually corrected ground truths. 
\end{itemize}

Our proposed method is implemented in pytorch. 
We use 32×H20 96GB GPUs to train our model with batch size 1 and the AdamW optimizer. In the reinforce learning stage, the learning rate starts from 1e-6 and decays to 0 following the liner schedule. During the rollout process, we sample 8 responses for each input prompt, with the KL divergence coefficient β set to 0.04.

\subsection{Comparisons on Medical VIE task}

\begin{table}[!h]
  \centering
  \caption{\label{citation-guide}
    \textbf{Performances on medical VIE task.} The field-level precision, recall, F1 scores and TED based accuracies are reported. For each metric, we bold the best results and underline the second-best results. Note that all VIE SFT and VIE RLVR models are 
     finetuned from Qwen2.5-VL-7B,  and VIE RLVR shows the best precision, recall, F1 scores. 
   \# In the column named 'output', 'OCR' means we use models to parse medical reports into markdown format and then apply GPT4o to extract JSON format answers, 'JSON' means we prompt models to directly output with JSON format. }
  \begin{tabular}{llccccc}
    \hline
    \textbf{Type}&\textbf{Model Name}&\#\textbf{Output}          & \textbf{F1}  & \textbf{Precision} & \textbf{Recall} & \textbf{TED Acc} \\
    \hline
    Pipeline&MinerU &OCR& 63.16 & 67.85 & 59.07 & 64.01 \\
    
    &Marker&OCR & 56.89 & 57.68 & 56.12 & 58.69 \\
    \hline
    
    Expert&GOT-OCR &OCR& 55.12 & 62.12 & 49.54 & 62.14 \\
    \hline 
    Commercial&TextIn&OCR&70.49&71.18&69.82&\textbf{77.28} \\
    \hline 
    
    General&GPT4o &JSON& 64.55 & 65.56 & 63.57 & 65.71 \\
    
    &InternVL-2.5-78B &JSON& 71.15 & 71.49 & 70.82 & \underline{74.42} \\
    &Qwen2.5-VL-72B &JSON&52.29 &66.92 &42.91 &65.14\\
    &Qwen2.5-VL-7B&JSON&67.61 &65.18 &70.23 &54.54\\
    \hline
    VIE SFT&JSON-SFT-100&JSON &66.69&63.91&69.71&55.21\\
    &OCR-SFT-17K&OCR & 66.94&63.25&71.08&57.08\\
    \hline
    VIE RLVR&OCR-SFT-17K-RL-100&JSON&76.23&76.18 & \underline{76.28}&72.72\\
    &RL-100(w/o sample)&JSON&\underline{77.10} &\underline{77.67} & \textbf{76.55}&69.64\\
    &RL-100&JSON &\textbf{77.81} &\textbf{79.85} & 75.88&68.49\\
    \hline
  \end{tabular}
  \label{tab:medical}
\end{table}

Our medical evaluation dataset consists of 203 medical report images uploaded by users, covering CT, MRI, X-ray, physical examination reports, endoscopy, prescriptions, urine tests, electrocardiograms, medical records, pathology, diagnostic tests, medicine boxes, blood tests, and ultrasound. These images include screenshots and scans, exhibiting diverse clarity levels and aspect ratios. To generate ground truth, under the guidance of doctors, we selected important fields from these medical images, used GPT-4o to extract values of these fields from the images, and manually corrected the answers. Finally, we obtained 203 image-JSON pairs as our test dataset.

The results of our models and baseline models are shown in Table~\ref{tab:medical}. We find that VIE RLVR models outperform all other models on the medical VIE task. They achieve SOTA performance in F1, precision, and recall metrics.
TextIn and InternVL-2.5-78B perform well on the TED accuracy metric. However, their other scores are relatively low. This indicates that these models can correctly extract the structure of medical reports but fail to parse text information accurately.
It is important to note that F1 score, precision, and recall are more critical than TED accuracy. TED accuracy only evaluates the tree topology extracted by the model, whereas F1 score, precision, and recall also assess the text information on each tree node. These metrics hold greater value in practical applications.

The substantial improvement of RLVR over SFT validates the effectiveness of our proposed reinforcement learning approach. 
Compared with SFT, RLVR is better suited to the unordered nature of JSON data, where a single image can correspond to multiple ground truths. In contrast, SFT uses cross-entropy loss on fixed JSON ground truths during training, which may cause data ambiguity and degrade model performance.
In addition, the model gains the ability to think and plan during the reinforcement learning process. This ability allows it to understand image structures more accurately and extract key textual information with higher precision.

For different VIE RLVR models, we first note that RL-100 maintains a 77.81 F1 score. It outperforms pipeline, expert, closed-source, general multimodal models, and VIE SFT models by nearly 10 points. This reveals that in domain-specific VIE tasks, leveraging high-quality small-scale datasets through RLVR enables significant performance gains for models.
Meanwhile, RL-100(w/o sample) achieves SOTA performance in the recall metric. The model attempts to extract as many fields as possible, but extracting redundant fields leads to a lower precision score.
Furthermore, OCR-SFT-17K-RL-100 reaches a higher TED accuracy. This means the model learns the tree topology of medical reports during the supervised finetuning stage.
To observe the OCR ability gained in the SFT stage, refer to Appendix \ref{sec:appendix_general_ocr}. Our model, trained on the OCR task with the medical report dataset, outperforms several pipeline and OCR expert models. Its overall score is close to the SOTA model with 72B parameters, demonstrating the high quality of our medical report dataset.

\subsection{The Impact of \colorbox{lightgray}{Think}}

\begin{table}[!h]
  \centering
  \caption{\label{citation-guide}
    \textbf{Impact of model thinking on medical VIE task.} The field-level precision, recall, F1 scores and TED based accuracies are reported. Note that both two models are 
     finetuned from Qwen2.5-VL-7B, the only difference is the thinking process in training and inferring stages.  }
  \begin{tabular}{lcccc}
    \hline
    \textbf{Model Name}          & \textbf{F1}  & \textbf{Precision} & \textbf{Recall} & \textbf{TED Accuracy} \\
    \hline
    
    RL-100 (w/o think)& 75.90 & 75.68 & 76.13 & 70.06 \\
    RL-100 (w/ think)&77.81 &79.85 & 75.88&68.49\\
    \hline
  \end{tabular}
  \label{tab:think}
\end{table}

To analyze the impact of model thinking, we compare the RL-100 model which is required to think while training and inferring, to the RL-100 model without thinking in training and inferring stages. Their performance on the medical VIE dataset are reported in Table~\ref{tab:think}. For the more considerable metrics, i.e. F1 scores that better reflects the model's VIE ability, model with thinking outperforms the model without thinking. In the thinking process, model understands the image better and plans to extract key information, thus thinking progress is important for MLLMs in VIE tasks. For case studies, refer to Appendix \ref{sec:appendix_case_think_nothink}.

\subsection{Analysis on General VIE tasks}

\begin{table}[htbp]
  \centering
  \caption{\textbf{Performances on general VIE tasks.} Various widely used benchmarks are chosen to evaluate the models general VIE performance. The field-level F1 scores and TED based accuracies are reported. For each metric, we bold the best results and underline the second-best results.}
  \label{tab:minimal_model_performance}
  \begin{tabular}{lcccccccc}
    \toprule
    & \multicolumn{2}{c}{\textbf{CORD}} & \multicolumn{2}{c}{\textbf{FUNSD}} & \multicolumn{2}{c}{\textbf{SROIE}}& \multicolumn{2}{c}{\textbf{Ticket}} \\
    \cmidrule(lr){2-3} \cmidrule(lr){4-5} \cmidrule(lr){6-7} \cmidrule(lr){8-9}
    \textbf{Model} & \textbf{TED Acc} & \textbf{F1} & \textbf{TED Acc} & \textbf{F1} & \textbf{TED Acc} & \textbf{F1}& \textbf{TED Acc} & \textbf{F1} \\
    \midrule
    Qwen2.5-VL-7B & 34.6 & 39.8 & 12.2 & 12.2 & \textbf{80.7} & \textbf{67.0} & 80.4 & \textbf{76.4} \\
    OCR-SFT-17K-RL-100 & \underline{59.6} & 46.8 & \underline{32.6} & \textbf{32.7} & 78.4 & \underline{66.4} & \underline{83.7} & 64.6 \\
    JSON-SFT-100 &36.1  & 40.8 & 14.5 & 14.5 & \underline{80.4} & \underline{66.4} & 72.8 & \underline{72.7} \\
    RL-100(w/o sample) & \textbf{68.7} & \textbf{49.2} & \textbf{34.0} & \underline{30.6} & 79.3 & 65.0 & \textbf{87.5} & 72.3 \\
    RL-100 & 45.7 & \underline{48.5} & 32.5 & 29.4 & 78.2 & 61.7 & 80.4 & 60.8 \\
    \bottomrule
  \end{tabular}
\end{table}

We also evaluate the medical VIE models trained with RLVR on general VIE tasks in order to demonstrate that VIE tasks exhibit strong domain-specific characteristics, VIE tasks in different domains vary significantly, making it challenging for models to acquire strong general VIE capabilities through training on specific-domain VIE tasks.
There are four widerly used VIE benchmarks: 

\begin{itemize}
\item CORD\cite{park2019cord}: The Consolidated Receipt Dataset (CORD) serves as a public benchmark comprising 800 training, 100 validation, and 100 test receipt images. The textual content of these receipts is encoded in the Latin alphabet. The dataset features 30 unique fields, including menu name, quantity, total price, and others. Notably, the information exhibits complex structures, such as nested groups and hierarchical organizations, which bears a resemblance to our medical VIE dataset.
\item FUNSD\cite{Jaume2019FUNSD}: FUNSD is a dataset for form understanding in noisy scanned documents, with 199 real, fully annotated forms for tasks like text detection and layout analysis. There are four fields to extracted, named 'question' 'answer' 'header' and 'others', and each field correspond to a list of values, which are similar to our medical VIE dataset.
\item SROIE\cite{Huang2019SROIE}: SROIE is the most widely adopted dataset, significantly advancing the field's development. The dataset comprises scanned English printed receipts, with each image accompanied by comprehensive OCR annotations and values for four key text fields.
\item Ticket\cite{guo2019eaten}: This public benchmark dataset comprises 1,500 training and 400 test images of Chinese train tickets. It includes eight fields, such as ticket number, departure station, and train number. The information structure is straightforward, with each key guaranteed to appear only once and the location of each field fixed.
\end{itemize}

From Table~\ref{tab:minimal_model_performance} we find that on CORD dataset, the VIE RLVR models outperform the base model with nearly 10 points in F1 score and 34 points in TED accuracy, and on FUNSD dataset VIE RLVR models outperform the base model with nearly 20 points in F1 score and 22 points in TED accuracy. However the performance of VIE RLVR models are poor on Ticket and  SROIE datasets. We note that images in CORD benchmark are receipts with 30 complex fields, having some sub-fields to extract, which is very similar to our medical VIE evaluation dataset, while the SROIE and Ticket datasets are composed of images with few information to extract. 
Additionally, the comparision of RL-100(w/o sample) with RL-100 in  Table~\ref{tab:minimal_model_performance}
is different from results in Table~\ref{tab:medical}
,  
indicating a significant gap exists among different VIE tasks.   

\section{Conclusion}

In this paper, we propose a novel method for training MLLMs for medical VIE tasks, leveraging the RLVR framework.
Our method applies merely 100 high quality samples for training, 
with our proposed reward mechanism being able to reduce model hallucination as well as ensure models to capture all the key fields. 
Extensive experiments demonstrate the effectiveness of our VIE RLVR models.
On medical report dataset, our models trained with merely 100 samples approach SOTA performance on the mainly considered metrics, such as F1, precision and recall,
exceeding commercial models like GPT-4o for more than 10 points,
showing the advantage of our proposed method. 
Results on different general VIE tasks also reveal the gap between different tasks.
Studies considering different data sampling strategy and thinking process are also given, validating the high efficiency of our method. 


\bibliography{custom}

\begin{thebibliography}{59}
\providecommand{\natexlab}[1]{#1}

\bibitem[{Adnan et~al.(2024)Adnan, Tang, Zouggari, Laatiri, Lam, and Caspani}]{adnan2024layoutlmv3}
Wiam Adnan, Joel Tang, Yassine Bel~Khayat Zouggari, Seif~Edinne Laatiri, Laurent Lam, and Fabien Caspani. 2024.
\newblock A layoutlmv3-based model for enhanced relation extraction in visually-rich documents.
\newblock In \emph{International Conference on Document Analysis and Recognition}, pages 160--174. Springer.

\bibitem[{Appalaraju et~al.(2021)Appalaraju, Jasani, Kota, Xie, and Manmatha}]{appalaraju2021docformer}
Srikar Appalaraju, Bhavan Jasani, Bhargava~Urala Kota, Yusheng Xie, and R~Manmatha. 2021.
\newblock Docformer: End-to-end transformer for document understanding.
\newblock In \emph{Proceedings of the IEEE/CVF international conference on computer vision}, pages 993--1003.

\bibitem[{Authors(2020)}]{paddleocr2020}
PaddlePaddle Authors. 2020.
\newblock Paddleocr, awesome multilingual ocr toolkits based on paddlepaddle.
\newblock \url{https://github.com/PaddlePaddle/PaddleOCR}.

\bibitem[{Bai et~al.(2025)Bai, Chen, Liu, Wang, Ge, Song, Dang, Wang, Wang, Tang et~al.}]{bai2025qwen2}
Shuai Bai, Keqin Chen, Xuejing Liu, Jialin Wang, Wenbin Ge, Sibo Song, Kai Dang, Peng Wang, Shijie Wang, Jun Tang, and 1 others. 2025.
\newblock Qwen2. 5-vl technical report.
\newblock \emph{arXiv preprint arXiv:2502.13923}.

\bibitem[{Cao et~al.(2022)Cao, Li, Ma, Jiang, Guo, Hu, Liu, Liu, and Ren}]{cao2022query}
Haoyu Cao, Xin Li, Jiefeng Ma, Deqiang Jiang, Antai Guo, Yiqing Hu, Hao Liu, Yinsong Liu, and Bo~Ren. 2022.
\newblock Query-driven generative network for document information extraction in the wild.
\newblock In \emph{Proceedings of the 30th ACM International Conference on Multimedia}, pages 4261--4271.

\bibitem[{Chen et~al.(2022)Chen, Lv, Cui, Zhang, and Wei}]{chen2022xdoc}
Jingye Chen, Tengchao Lv, Lei Cui, Cha Zhang, and Furu Wei. 2022.
\newblock Xdoc: Unified pre-training for cross-format document understanding.
\newblock \emph{arXiv preprint arXiv:2210.02849}.

\bibitem[{Chen et~al.(2025)Chen, Wang, Cao, Liu, Gao, Cui, Zhu, Ye, Tian, Liu, Gu, Wang, Li, Ren, Chen, Luo, Wang, Jiang, Wang, He, Shi, Zhang, Lv, Wang, Shao, Chu, Tu, He, Wu, Deng, Ge, Chen, Zhang, Wang, Dou, Lu, Zhu, Lu, Lin, Qiao, Dai, and Wang}]{chen2025expandingperformanceboundariesopensource}
Zhe Chen, Weiyun Wang, Yue Cao, Yangzhou Liu, Zhangwei Gao, Erfei Cui, Jinguo Zhu, Shenglong Ye, Hao Tian, Zhaoyang Liu, Lixin Gu, Xuehui Wang, Qingyun Li, Yimin Ren, Zixuan Chen, Jiapeng Luo, Jiahao Wang, Tan Jiang, Bo~Wang, and 23 others. 2025.
\newblock \href {https://arxiv.org/abs/2412.05271} {Expanding performance boundaries of open-source multimodal models with model, data, and test-time scaling}.
\newblock \emph{Preprint}, arXiv:2412.05271.

\bibitem[{Duan et~al.(2025)Duan, Fang, Wang, Wang, Huang, Zeng, Li, and Liu}]{duan2025got}
Chengqi Duan, Rongyao Fang, Yuqing Wang, Kun Wang, Linjiang Huang, Xingyu Zeng, Hongsheng Li, and Xihui Liu. 2025.
\newblock Got-r1: Unleashing reasoning capability of mllm for visual generation with reinforcement learning.
\newblock \emph{arXiv preprint arXiv:2505.17022}.

\bibitem[{Feng et~al.(2025)Feng, Wei, Fei, Shi, Han, Liao, Lu, Wu, Liu, Lin et~al.}]{feng2025dolphin}
Hao Feng, Shu Wei, Xiang Fei, Wei Shi, Yingdong Han, Lei Liao, Jinghui Lu, Binghong Wu, Qi~Liu, Chunhui Lin, and 1 others. 2025.
\newblock Dolphin: Document image parsing via heterogeneous anchor prompting.
\newblock \emph{arXiv preprint arXiv:2505.14059}.

\bibitem[{Guo et~al.(2025)Guo, Yang, Zhang, Song, Zhang, Xu, Zhu, Ma, Wang, Bi et~al.}]{guo2025deepseek}
Daya Guo, Dejian Yang, Haowei Zhang, Junxiao Song, Ruoyu Zhang, Runxin Xu, Qihao Zhu, Shirong Ma, Peiyi Wang, Xiao Bi, and 1 others. 2025.
\newblock Deepseek-r1: Incentivizing reasoning capability in llms via reinforcement learning.
\newblock \emph{arXiv preprint arXiv:2501.12948}.

\bibitem[{Guo et~al.(2019)Guo, Qin, Liu, Han, Liu, and Ding}]{guo2019eaten}
He~Guo, Xiameng Qin, Jiaming Liu, Junyu Han, Jingtuo Liu, and Errui Ding. 2019.
\newblock Eaten: Entity-aware attention for single shot visual text extraction.
\newblock In \emph{2019 International Conference on Document Analysis and Recognition (ICDAR)}, pages 254--259. IEEE.

\bibitem[{Hong et~al.(2022)Hong, Kim, Ji, Hwang, Nam, and Park}]{hong2022bros}
Teakgyu Hong, Donghyun Kim, Mingi Ji, Wonseok Hwang, Daehyun Nam, and Sungrae Park. 2022.
\newblock Bros: A pre-trained language model focusing on text and layout for better key information extraction from documents.
\newblock In \emph{Proceedings of the AAAI Conference on Artificial Intelligence}, volume~36, pages 10767--10775.

\bibitem[{Huang et~al.(2025)Huang, Jia, Zhai, Cao, Ye, Zhao, Xu, Hu, and Lin}]{huang2025vision}
Wenxuan Huang, Bohan Jia, Zijie Zhai, Shaosheng Cao, Zheyu Ye, Fei Zhao, Zhe Xu, Yao Hu, and Shaohui Lin. 2025.
\newblock Vision-r1: Incentivizing reasoning capability in multimodal large language models.
\newblock \emph{arXiv preprint arXiv:2503.06749}.

\bibitem[{Huang et~al.(2022)Huang, Lv, Cui, Lu, and Wei}]{huang2022layoutlmv3}
Yupan Huang, Tengchao Lv, Lei Cui, Yutong Lu, and Furu Wei. 2022.
\newblock Layoutlmv3: Pre-training for document ai with unified text and image masking.
\newblock In \emph{Proceedings of the 30th ACM international conference on multimedia}, pages 4083--4091.

\bibitem[{Huang et~al.(2019{\natexlab{a}})Huang, Chen, He, Bai, Karatzas, Lu, and Jawahar}]{Huang2019SROIE}
Zheng Huang, Kai Chen, Jianhua He, Xiang Bai, Dimosthenis Karatzas, Shijian Lu, and C.V. Jawahar. 2019{\natexlab{a}}.
\newblock \href {https://doi.org/10.1109/ICDAR.2019.00244} {{ICDAR} 2019 competition on scanned receipt {OCR} and information extraction ({SROIE})}.
\newblock In \emph{2019 International Conference on Document Analysis and Recognition ({ICDAR})}, pages 1516--1520.

\bibitem[{Huang et~al.(2019{\natexlab{b}})Huang, Chen, He, Bai, Karatzas, Lu, and Jawahar}]{huang2019icdar2019}
Zheng Huang, Kai Chen, Jianhua He, Xiang Bai, Dimosthenis Karatzas, Shijian Lu, and CV~Jawahar. 2019{\natexlab{b}}.
\newblock Icdar2019 competition on scanned receipt ocr and information extraction.
\newblock In \emph{2019 International Conference on Document Analysis and Recognition (ICDAR)}, pages 1516--1520. IEEE.

\bibitem[{Hui et~al.(2024)Hui, Yang, Cui, Yang, Liu, Zhang, Liu, Zhang, Yu, Lu et~al.}]{hui2024qwen2}
Binyuan Hui, Jian Yang, Zeyu Cui, Jiaxi Yang, Dayiheng Liu, Lei Zhang, Tianyu Liu, Jiajun Zhang, Bowen Yu, Keming Lu, and 1 others. 2024.
\newblock Qwen2. 5-coder technical report.
\newblock \emph{arXiv preprint arXiv:2409.12186}.

\bibitem[{Jaech et~al.(2024)Jaech, Kalai, Lerer, Richardson, El-Kishky, Low, Helyar, Madry, Beutel, Carney et~al.}]{jaech2024openai}
Aaron Jaech, Adam Kalai, Adam Lerer, Adam Richardson, Ahmed El-Kishky, Aiden Low, Alec Helyar, Aleksander Madry, Alex Beutel, Alex Carney, and 1 others. 2024.
\newblock Openai o1 system card.
\newblock \emph{arXiv preprint arXiv:2412.16720}.

\bibitem[{Jaume et~al.(2019)Jaume, Ekenel, and Thiran}]{Jaume2019FUNSD}
Guillaume Jaume, Haz{\i}m~Kemal Ekenel, and Jean{-}Philippe Thiran. 2019.
\newblock \href {https://doi.org/10.48550/ARXIV.1905.13538} {{FUNSD}: A dataset for form understanding in noisy scanned documents}.
\newblock \emph{arXiv preprint arXiv:1905.13538}.

\bibitem[{Kim et~al.(2022)Kim, Hong, Yim, Nam, Park, Yim, Hwang, Yun, Han, and Park}]{kim2022ocr}
Geewook Kim, Teakgyu Hong, Moonbin Yim, JeongYeon Nam, Jinyoung Park, Jinyeong Yim, Wonseok Hwang, Sangdoo Yun, Dongyoon Han, and Seunghyun Park. 2022.
\newblock Ocr-free document understanding transformer.
\newblock In \emph{European Conference on Computer Vision}, pages 498--517. Springer.

\bibitem[{Kim et~al.(2021)Kim, Hong, Yim, Park, Yim, Hwang, Yun, Han, and Park}]{kim2021donut}
Geewook Kim, Teakgyu Hong, Moonbin Yim, Jinyoung Park, Jinyeong Yim, Wonseok Hwang, Sangdoo Yun, Dongyoon Han, and Seunghyun Park. 2021.
\newblock Donut: Document understanding transformer without ocr.
\newblock \emph{arXiv preprint arXiv:2111.15664}, 7(15):2.

\bibitem[{Kuang et~al.(2023)Kuang, Hua, Liang, Yang, Jiang, Ren, and Bai}]{kuang2023visual}
Jianfeng Kuang, Wei Hua, Dingkang Liang, Mingkun Yang, Deqiang Jiang, Bo~Ren, and Xiang Bai. 2023.
\newblock Visual information extraction in the wild: practical dataset and end-to-end solution.
\newblock In \emph{International Conference on Document Analysis and Recognition}, pages 36--53. Springer.

\bibitem[{Li et~al.(2024)Li, Wei, Chen, Li, Tao, and Xu}]{li2024improving}
Yiming Li, Qiang Wei, Xinghan Chen, Jianfu Li, Cui Tao, and Hua Xu. 2024.
\newblock Improving tabular data extraction in scanned laboratory reports using deep learning models.
\newblock \emph{Journal of Biomedical Informatics}, 159:104735.

\bibitem[{Liu et~al.(2025{\natexlab{a}})Liu, Peng, Zhong, Yue, Lu, Yu, and Jia}]{liu2025seg}
Yuqi Liu, Bohao Peng, Zhisheng Zhong, Zihao Yue, Fanbin Lu, Bei Yu, and Jiaya Jia. 2025{\natexlab{a}}.
\newblock Seg-zero: Reasoning-chain guided segmentation via cognitive reinforcement.
\newblock \emph{arXiv preprint arXiv:2503.06520}.

\bibitem[{Liu et~al.(2025{\natexlab{b}})Liu, Tu, Ye, Xiao, Zhang, Cui, Zhu, Ju, Li, and Xie}]{liu2025exploring}
Zhaocheng Liu, Quan Tu, Wen Ye, Yu~Xiao, Zhishou Zhang, Hengfu Cui, Yalun Zhu, Qiang Ju, Shizheng Li, and Jian Xie. 2025{\natexlab{b}}.
\newblock Exploring the inquiry-diagnosis relationship with advanced patient simulators.
\newblock \emph{arXiv preprint arXiv:2501.09484}.

\bibitem[{Liu et~al.(2025{\natexlab{c}})Liu, Sun, Zang, Dong, Cao, Duan, Lin, and Wang}]{liu2025visual}
Ziyu Liu, Zeyi Sun, Yuhang Zang, Xiaoyi Dong, Yuhang Cao, Haodong Duan, Dahua Lin, and Jiaqi Wang. 2025{\natexlab{c}}.
\newblock Visual-rft: Visual reinforcement fine-tuning.
\newblock \emph{arXiv preprint arXiv:2503.01785}.

\bibitem[{Luo et~al.(2023)Luo, Cheng, Zheng, and Yao}]{luo2023geolayoutlm}
Chuwei Luo, Changxu Cheng, Qi~Zheng, and Cong Yao. 2023.
\newblock Geolayoutlm: Geometric pre-training for visual information extraction.
\newblock In \emph{Proceedings of the IEEE/CVF conference on computer vision and pattern recognition}, pages 7092--7101.

\bibitem[{Ma et~al.(2023)Ma, Gao, Zhang, Shang, Jin, Liu, Lv, Ni, Han, and Zong}]{ma2023extracting}
Ming-Wei Ma, Xian-Shu Gao, Ze-Yu Zhang, Shi-Yu Shang, Ling Jin, Pei-Lin Liu, Feng Lv, Wei Ni, Yu-Chen Han, and Hui Zong. 2023.
\newblock Extracting laboratory test information from paper-based reports.
\newblock \emph{BMC Medical Informatics and Decision Making}, 23(1):251.

\bibitem[{OpenAI(2024{\natexlab{a}})}]{openai4o}
OpenAI. 2024{\natexlab{a}}.
\newblock {GPT-4o system card.}
\newblock \url{https://openai.com/index/gpt-4o-system-card/}.

\bibitem[{OpenAI(2024{\natexlab{b}})}]{openaio1}
OpenAI. 2024{\natexlab{b}}.
\newblock {Openai o1 system card.}
\newblock \url{https://openai.com/index/openai-o1-system-card/}.

\bibitem[{Ouyang et~al.(2025)Ouyang, Qu, Zhou, Zhu, Zhang, Lin, Wang, Zhao, Jiang, Zhao et~al.}]{ouyang2025omnidocbench}
Linke Ouyang, Yuan Qu, Hongbin Zhou, Jiawei Zhu, Rui Zhang, Qunshu Lin, Bin Wang, Zhiyuan Zhao, Man Jiang, Xiaomeng Zhao, and 1 others. 2025.
\newblock Omnidocbench: Benchmarking diverse pdf document parsing with comprehensive annotations.
\newblock In \emph{Proceedings of the Computer Vision and Pattern Recognition Conference}, pages 24838--24848.

\bibitem[{Pang(2024)}]{Pang_Faster_LayoutReader_based_2024}
Hantian Pang. 2024.
\newblock \href {https://github.com/ppaanngggg/layoutreader} {{Faster LayoutReader based on LayoutLMv3}}.

\bibitem[{Park et~al.(2019)Park, Shin, Lee, Lee, Surh, Seo, and Lee}]{park2019cord}
Seunghyun Park, Seung Shin, Bado Lee, Junyeop Lee, Jaeheung Surh, Minjoon Seo, and Hwalsuk Lee. 2019.
\newblock Cord: A consolidated receipt dataset for post-ocr parsing.
\newblock In \emph{Document Intelligence Workshop at NeurIPS 2019}.

\bibitem[{Powalski et~al.(2021)Powalski, Borchmann, Jurkiewicz, Dwojak, Pietruszka, and Pa{\l}ka}]{powalski2021going}
Rafa{\l} Powalski, {\L}ukasz Borchmann, Dawid Jurkiewicz, Tomasz Dwojak, Micha{\l} Pietruszka, and Gabriela Pa{\l}ka. 2021.
\newblock Going full-tilt boogie on document understanding with text-image-layout transformer.
\newblock In \emph{Document Analysis and Recognition--ICDAR 2021: 16th International Conference, Lausanne, Switzerland, September 5--10, 2021, Proceedings, Part II 16}, pages 732--747. Springer.

\bibitem[{Poznanski et~al.(2025)Poznanski, Borchardt, Dunkelberger, Huff, Lin, Rangapur, Wilhelm, Lo, and Soldaini}]{poznanski2025olmocr}
Jake Poznanski, Jon Borchardt, Jason Dunkelberger, Regan Huff, Daniel Lin, Aman Rangapur, Christopher Wilhelm, Kyle Lo, and Luca Soldaini. 2025.
\newblock olmocr: Unlocking trillions of tokens in pdfs with vision language models.
\newblock \emph{arXiv preprint arXiv:2502.18443}.

\bibitem[{Shao et~al.(2024)Shao, Wang, Zhu, Xu, Song, Bi, Zhang, Zhang, Li, Wu et~al.}]{shao2024deepseekmath}
Zhihong Shao, Peiyi Wang, Qihao Zhu, Runxin Xu, Junxiao Song, Xiao Bi, Haowei Zhang, Mingchuan Zhang, YK~Li, Y~Wu, and 1 others. 2024.
\newblock Deepseekmath: Pushing the limits of mathematical reasoning in open language models.
\newblock \emph{arXiv preprint arXiv:2402.03300}.

\bibitem[{Shen et~al.(2025)Shen, Liu, Li, Fang, Ma, Liao, Shen, Zhang, Zhao, Zhang et~al.}]{shen2025vlm}
Haozhan Shen, Peng Liu, Jingcheng Li, Chunxin Fang, Yibo Ma, Jiajia Liao, Qiaoli Shen, Zilun Zhang, Kangjia Zhao, Qianqian Zhang, and 1 others. 2025.
\newblock Vlm-r1: A stable and generalizable r1-style large vision-language model.
\newblock \emph{arXiv preprint arXiv:2504.07615}.

\bibitem[{Team et~al.(2025)Team, Du, Gao, Xing, Jiang, Chen, Li, Xiao, Du, Liao et~al.}]{team2025kimi}
Kimi Team, Angang Du, Bofei Gao, Bowei Xing, Changjiu Jiang, Cheng Chen, Cheng Li, Chenjun Xiao, Chenzhuang Du, Chonghua Liao, and 1 others. 2025.
\newblock Kimi k1. 5: Scaling reinforcement learning with llms.
\newblock \emph{arXiv preprint arXiv:2501.12599}.

\bibitem[{Wan et~al.(2024)Wan, Song, Yu, Liu, Cheng, Huang, Bai, Yao, and Yang}]{wan2024omniparser}
Jianqiang Wan, Sibo Song, Wenwen Yu, Yuliang Liu, Wenqing Cheng, Fei Huang, Xiang Bai, Cong Yao, and Zhibo Yang. 2024.
\newblock Omniparser: A unified framework for text spotting key information extraction and table recognition.
\newblock In \emph{Proceedings of the IEEE/CVF Conference on Computer Vision and Pattern Recognition}, pages 15641--15653.

\bibitem[{Wang et~al.(2024{\natexlab{a}})Wang, Gu, Liang, Xu, Zhang, Shi, and He}]{wang2024unimernetuniversalnetworkrealworld}
Bin Wang, Zhuangcheng Gu, Guang Liang, Chao Xu, Bo~Zhang, Botian Shi, and Conghui He. 2024{\natexlab{a}}.
\newblock \href {https://arxiv.org/abs/2404.15254} {Unimernet: A universal network for real-world mathematical expression recognition}.
\newblock \emph{Preprint}, arXiv:2404.15254.

\bibitem[{Wang et~al.(2024{\natexlab{b}})Wang, Xu, Zhao, Ouyang, Wu, Zhao, Xu, Liu, Qu, Shang et~al.}]{wang2024mineru}
Bin Wang, Chao Xu, Xiaomeng Zhao, Linke Ouyang, Fan Wu, Zhiyuan Zhao, Rui Xu, Kaiwen Liu, Yuan Qu, Fukai Shang, and 1 others. 2024{\natexlab{b}}.
\newblock Mineru: An open-source solution for precise document content extraction.
\newblock \emph{arXiv preprint arXiv:2409.18839}.

\bibitem[{Wang et~al.()Wang, Zheng, Wu, Li, Li, and Zhang}]{wangchatschema}
Fei Wang, Yuewen Zheng, Jingyi Wu, Qing Li, Pengfei Li, and Luxia Zhang.
\newblock Chatschema: development and validation of a pipeline for extracting structured nursing information with large multimodal models.
\newblock \emph{Interdisciplinary Nursing Research}, pages 10--1097.

\bibitem[{Wang et~al.(2021{\natexlab{a}})Wang, Liu, Jin, Tang, Zhang, Zhang, Wang, Wu, and Cai}]{wang2021towards}
Jiapeng Wang, Chongyu Liu, Lianwen Jin, Guozhi Tang, Jiaxin Zhang, Shuaitao Zhang, Qianying Wang, Yaqiang Wu, and Mingxiang Cai. 2021{\natexlab{a}}.
\newblock Towards robust visual information extraction in real world: new dataset and novel solution.
\newblock In \emph{Proceedings of the AAAI Conference on Artificial Intelligence}, volume~35, pages 2738--2745.

\bibitem[{Wang et~al.(2021{\natexlab{b}})Wang, Zhang, Qi, Liu, Zhang, Lyu, Han, Liu, Ding, and Shi}]{wang2021pgnet}
Pengfei Wang, Chengquan Zhang, Fei Qi, Shanshan Liu, Xiaoqiang Zhang, Pengyuan Lyu, Junyu Han, Jingtuo Liu, Errui Ding, and Guangming Shi. 2021{\natexlab{b}}.
\newblock Pgnet: Real-time arbitrarily-shaped text spotting with point gathering network.
\newblock In \emph{Proceedings of the AAAI Conference on Artificial Intelligence}, volume~35, pages 2782--2790.

\bibitem[{Wei et~al.(2024)Wei, Liu, Chen, Wang, Kong, Xu, Ge, Zhao, Sun, Peng et~al.}]{wei2024general}
Haoran Wei, Chenglong Liu, Jinyue Chen, Jia Wang, Lingyu Kong, Yanming Xu, Zheng Ge, Liang Zhao, Jianjian Sun, Yuang Peng, and 1 others. 2024.
\newblock General ocr theory: Towards ocr-2.0 via a unified end-to-end model.

\bibitem[{Xia et~al.(2024)Xia, Mao, Yan, Zhou, Zhang, Peng, Pi, Fu, Wu, Ye, Feng, Wang, Xu, He, Cai, Dou, Shi, Zhou, Wang, Wang, Yan, Wu, and Qiao}]{xia2024docgenomeopenlargescalescientific}
Renqiu Xia, Song Mao, Xiangchao Yan, Hongbin Zhou, Bo~Zhang, Haoyang Peng, Jiahao Pi, Daocheng Fu, Wenjie Wu, Hancheng Ye, Shiyang Feng, Bin Wang, Chao Xu, Conghui He, Pinlong Cai, Min Dou, Botian Shi, Sheng Zhou, Yongwei Wang, and 4 others. 2024.
\newblock \href {https://arxiv.org/abs/2406.11633} {Docgenome: An open large-scale scientific document benchmark for training and testing multi-modal large language models}.
\newblock \emph{Preprint}, arXiv:2406.11633.

\bibitem[{Xu et~al.(2024)Xu, Jin, Hao, Song, Sun, and Yuan}]{xu2024llava}
Guowei Xu, Peng Jin, Li~Hao, Yibing Song, Lichao Sun, and Li~Yuan. 2024.
\newblock Llava-o1: Let vision language models reason step-by-step.
\newblock \emph{arXiv preprint arXiv:2411.10440}.

\bibitem[{Xu et~al.(2020{\natexlab{a}})Xu, Xu, Lv, Cui, Wei, Wang, Lu, Florencio, Zhang, Che et~al.}]{xu2020layoutlmv2}
Yang Xu, Yiheng Xu, Tengchao Lv, Lei Cui, Furu Wei, Guoxin Wang, Yijuan Lu, Dinei Florencio, Cha Zhang, Wanxiang Che, and 1 others. 2020{\natexlab{a}}.
\newblock Layoutlmv2: Multi-modal pre-training for visually-rich document understanding.
\newblock \emph{arXiv preprint arXiv:2012.14740}.

\bibitem[{Xu et~al.(2020{\natexlab{b}})Xu, Li, Cui, Huang, Wei, and Zhou}]{xu2020layoutlm}
Yiheng Xu, Minghao Li, Lei Cui, Shaohan Huang, Furu Wei, and Ming Zhou. 2020{\natexlab{b}}.
\newblock Layoutlm: Pre-training of text and layout for document image understanding.
\newblock In \emph{Proceedings of the 26th ACM SIGKDD international conference on knowledge discovery \& data mining}, pages 1192--1200.

\bibitem[{Yang et~al.(2025)Yang, He, Pan, Jiang, Deng, Yang, Lu, Yin, Rao, Zhu et~al.}]{yang2025r1}
Yi~Yang, Xiaoxuan He, Hongkun Pan, Xiyan Jiang, Yan Deng, Xingtao Yang, Haoyu Lu, Dacheng Yin, Fengyun Rao, Minfeng Zhu, and 1 others. 2025.
\newblock R1-onevision: Advancing generalized multimodal reasoning through cross-modal formalization.
\newblock \emph{arXiv preprint arXiv:2503.10615}.

\bibitem[{Yao et~al.(2024)Yao, Zhuang, Li, and Wei}]{yao2024learning}
Minghong Yao, Liansheng Zhuang, Houqiang Li, and Jiuchang Wei. 2024.
\newblock Learning label dependencies for visual information extraction.
\newblock In \emph{Proceedings of the Thirty-Third International Joint Conference on Artificial Intelligence}, pages 6615--6623.

\bibitem[{Ying et~al.(2024)Ying, Zhang, Li, Zhou, Shao, Fei, Ma, Hong, Liu, Wang et~al.}]{ying2024internlm}
Huaiyuan Ying, Shuo Zhang, Linyang Li, Zhejian Zhou, Yunfan Shao, Zhaoye Fei, Yichuan Ma, Jiawei Hong, Kuikun Liu, Ziyi Wang, and 1 others. 2024.
\newblock Internlm-math: Open math large language models toward verifiable reasoning.
\newblock \emph{arXiv preprint arXiv:2402.06332}.

\bibitem[{Yu et~al.(2025{\natexlab{a}})Yu, Lin, Zhao, Yin, Wei, Peng, Wei, Sun, Han, Ge et~al.}]{yu2025perception}
En~Yu, Kangheng Lin, Liang Zhao, Jisheng Yin, Yana Wei, Yuang Peng, Haoran Wei, Jianjian Sun, Chunrui Han, Zheng Ge, and 1 others. 2025{\natexlab{a}}.
\newblock Perception-r1: Pioneering perception policy with reinforcement learning.
\newblock \emph{arXiv preprint arXiv:2504.07954}.

\bibitem[{Yu et~al.(2025{\natexlab{b}})Yu, Zhang, Zhu, Yuan, Zuo, Yue, Fan, Liu, Liu, Liu et~al.}]{yu2025dapo}
Qiying Yu, Zheng Zhang, Ruofei Zhu, Yufeng Yuan, Xiaochen Zuo, Yu~Yue, Tiantian Fan, Gaohong Liu, Lingjun Liu, Xin Liu, and 1 others. 2025{\natexlab{b}}.
\newblock Dapo: An open-source llm reinforcement learning system at scale.
\newblock \emph{arXiv preprint arXiv:2503.14476}.

\bibitem[{Yu et~al.(2021)Yu, Lu, Qi, Gong, and Xiao}]{yu2021pick}
Wenwen Yu, Ning Lu, Xianbiao Qi, Ping Gong, and Rong Xiao. 2021.
\newblock Pick: processing key information extraction from documents using improved graph learning-convolutional networks.
\newblock In \emph{2020 25th International conference on pattern recognition (ICPR)}, pages 4363--4370. IEEE.

\bibitem[{Zhang et~al.(2020)Zhang, Xu, Cheng, Pu, Lu, Qiao, Niu, and Wu}]{zhang2020trie}
Peng Zhang, Yunlu Xu, Zhanzhan Cheng, Shiliang Pu, Jing Lu, Liang Qiao, Yi~Niu, and Fei Wu. 2020.
\newblock Trie: end-to-end text reading and information extraction for document understanding.
\newblock In \emph{Proceedings of the 28th ACM International Conference on Multimedia}, pages 1413--1422.

\bibitem[{Zhao et~al.(2024)Zhao, Kang, Wang, and He}]{zhao2024doclayoutyoloenhancingdocumentlayout}
Zhiyuan Zhao, Hengrui Kang, Bin Wang, and Conghui He. 2024.
\newblock \href {https://arxiv.org/abs/2410.12628} {Doclayout-yolo: Enhancing document layout analysis through diverse synthetic data and global-to-local adaptive perception}.
\newblock \emph{Preprint}, arXiv:2410.12628.

\bibitem[{Zheng et~al.(2022)Zheng, Liu, Sun, and He}]{zheng2022improving}
Lianchi Zheng, Xiaoming Liu, Zhihui Sun, and Yuxiang He. 2022.
\newblock Improving medical ocr information extraction with integrated bert and layoutxlm models.
\newblock In \emph{China Health Information Processing Conference}, pages 164--173. Springer.

\bibitem[{Zhou et~al.(2025)Zhou, Li, Wang, Cheng, Zhou, and Hsieh}]{zhou2503r1}
Hengguang Zhou, Xirui Li, Ruochen Wang, Minhao Cheng, Tianyi Zhou, and Cho-Jui Hsieh. 2025.
\newblock R1-zero’s" aha moment" in visual reasoning on a 2b non-sft model, 2025.
\newblock \emph{URL https://arxiv. org/abs/2503.05132}.

\end{thebibliography}

\appendix

\section{Predefined Medical Schema}
\label{sec:appendix_pms} 
\begin{tcolorbox}[left=0mm,right=0mm,top=0mm,bottom=0mm,boxsep=1mm,arc=0mm,boxrule=0pt, frame empty, breakable]

\begin{lstlisting}
{
    "Name": "",  // Patient's name, output as empty if not available
    "Gender": "",  // Patient's gender, output as empty if not available
    "Age": "",  // Patient's age, output as empty if not available
    "Examination Time": "",  // Time when the patient was examined, output as empty if not available
    "Department": "",  // Department where the patient was examined, output as empty if not available
    "Examination Name": "",  // Name of the examination performed on the patient, fill in if listed separately, output as empty if not available, do not extract from indicators or examination descriptions
    "Examination Site": "",  // Site where the patient was examined, fill in if listed separately, output as empty if not available, do not extract from indicators or examination descriptions
    "Indicators": [  // Various indicators of the examination items for the patient, generally displayed in a table
        {
            "Item Name": "",  
            "Result": "",  
            "Unit": "",  
            "Reference Range": "", 
            "Abnormal Mark": "",  
            "Detection Method": "",  
            "Result Status": "", 
            "Clinical Indication": "",  
            "Critical Value": "" 
        }
    ],
    "Examination Description": "",  // Description of the examination results, such as imaging findings, ultrasound findings, specimen descriptions, gross findings, microscopic findings, pathological descriptions, etc., output as empty if not available
    "Diagnosis": "",  // Preliminary/clinical/pathological diagnosis of the examination results; and identify which type of diagnosis it is. Output as empty if not available
    "Treatment Recommendations": "",  // Treatment recommendations given in the report, fill in if listed separately, output as empty if not available, do not extract from examination descriptions or diagnoses
    "Sample Collection Time": "",  // Time when the examination sample was collected, output as empty if not available
    "Others": ""  // Information in the report that affects the interpretation of the medical report but is not included in the above fields, output as empty if not available
}
\end{lstlisting}
\end{tcolorbox}

\section{Analysis on General OCR tasks}
\label{sec:appendix_general_ocr} 

\begin{table}[!htbp]
  \centering
  \caption{\textbf{Performances on general OCR task.} Results of different models on OmniDocBench\cite{ouyang2025omnidocbench} are reported, the scores are normalized edit distances. For each metric, we bold the best results and underline the second-best results.}
  \scalebox{0.8}{
  \begin{tabular}{l|ccccccccc|c}
    \hline
    \textbf{Models} & \textbf{Book} & \textbf{Slides} & \textbf{Financial}& \textbf{Textbook} & \textbf{Exam} & \textbf{Magazine} & \textbf{Academic} & \textbf{Notes} & \textbf{News} & \textbf{Overall} \\
    \hline
    MinerU  & \textbf{0.055} & 0.124 & \textbf{0.033} & \underline{0.102} & \textbf{0.159} & \underline{0.072} & \textbf{0.025} & 0.984 & \textbf{0.171} & 0.206 \\
    \hline
     Marker & \underline{0.074} & 0.34 & 0.089 & 0.319 & 0.452 & 0.153 & \underline{0.059} & 0.651 & \underline{0.192} & 0.274 \\
    \hline
     Mathpix & 0.131 & 0.22 & 0.202 & 0.216 & 0.278 & 0.147 & 0.091 & 0.634 & 0.69 & 0.3 \\
    \hline
    GOT-OCR  & 0.111 & 0.222 & 0.067 & 0.132 & 0.204 & 0.198 & 0.179 & 0.388 & 0.771 & 0.267 \\
    \hline
     Nougat & 0.734 & 0.958 & 1.000 & 0.820 & 0.930 & 0.83 & 0.214 & 0.991 & 0.871 & 0.806 \\
    \hline
     GPT4o  & 0.157 & 0.163 & 0.348 & 0.187 & 0.281 & 0.173 & 0.146 & 0.607 & 0.751 & 0.316 \\
    \hline
     Qwen2-VL-72B  & 0.096 & \textbf{0.061} & \underline{0.047} & 0.149 & 0.195 & \textbf{0.071} & 0.085 & \textbf{0.168} & 0.676 & \textbf{0.179} \\
    \hline
     InternVL2-76B  & 0.216 & 0.098 & 0.162 & 0.184 & 0.247 & 0.150 & 0.419 & 0.226 & 0.903 & 0.3 \\
    \hline
     Qwen2.5-VL-7B  & 0.2225 & 0.5704 & 0.1094 & 0.2545 & 0.1897 & 0.3580 & 0.2419 & 0.4176 & 0.6975 & 0.3591 \\
    \hline
    OCR-SFT-17K  & 0.1813 & \underline{0.0878} & 0.0915 & \textbf{0.0943} & \underline{0.1612} & 0.1504 & 0.1599 & \underline{0.1778} & 0.6334 & \underline{0.1984} \\
    \hline
  \end{tabular}
  }
  \label{tab:omnidocbench}
\end{table}

In Tab~\ref{tab:omnidocbench} we compare our model OCR-SFT-17K trained for OCR task using our 17K medical report dataset, with other widely considered pipeline models, expert models, general MLLMs as in \ref{tab:minimal_model_performance}. On the overall metric our model outperform GPT4o and many other models, reached the second place and very close to the SOTA performance, which means the medical report dataset is high quality and high diversity. 

\section{Case study of model with and without thinking}
\label{sec:appendix_case_think_nothink} 

We give study of two cases to analyze the effect of model's thinking progress, refering to Tab \ref{tab:thinkcase1} and Tab 6.

\begin{table}[!h]
  \centering
  \begin{center}
  
    \includegraphics[width=0.6\textwidth, keepaspectratio=true]{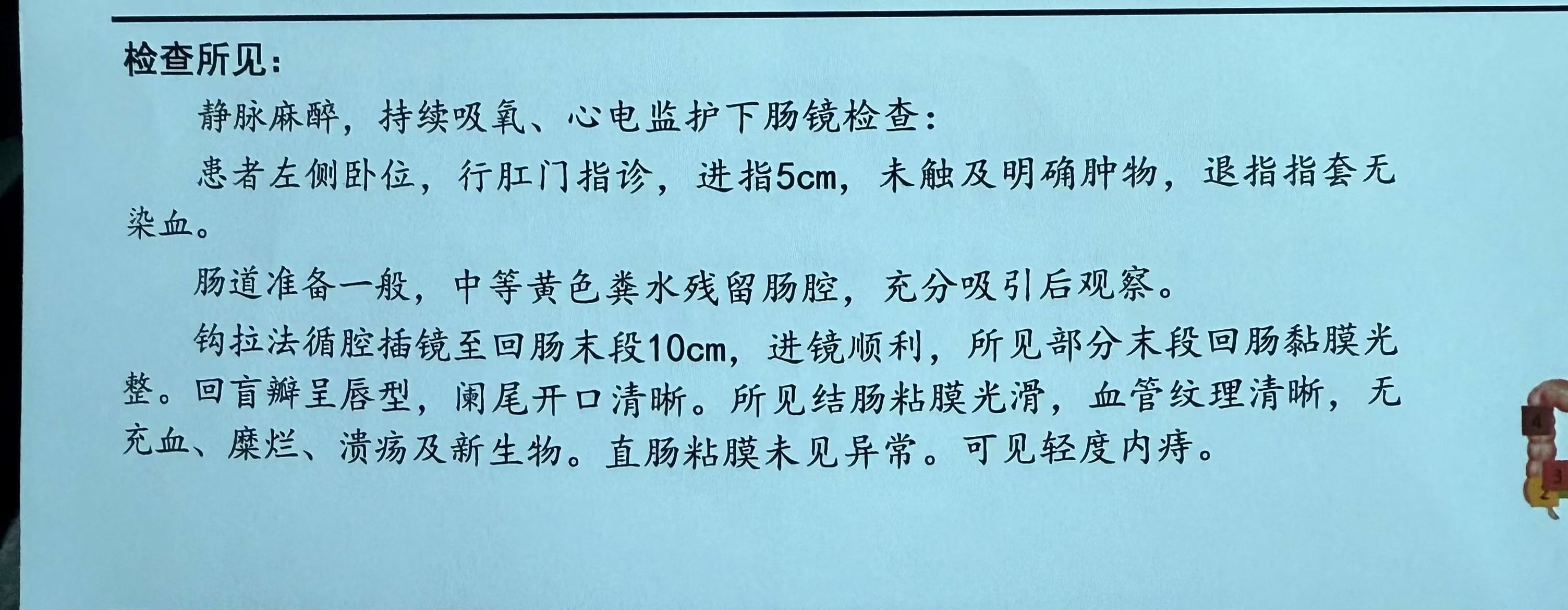}  
    
    \vspace{2pt}
    

  \end{center}
  \scalebox{0.7}{
  \begin{tabularx}{\textwidth}{X|X|X}
    \hline
    &\textbf{RL-100 (w/ think)} & \textbf{RL-100 (w/o think)}  \\
    \hline 
    Model Output&<think>
从图片中可以看到，这是一份肠镜检查报告。\textcolor{blue}{报告中没有明确提到患者的姓名、性别、年龄、检查时间、科室、检查名称、检查部位等信息。因此，这些信息无法从图片中提取。}报告中主要描述了肠镜检查的过程和结果，包括检查部位、检查描述、诊断等信息。根据这些信息，可以构建出一个符合要求的JSON格式的输出。
</think>
<answer>
```json
\{
    "姓名": "",
    "性别": "",
    "年龄": "",
    "检查时间": "",
    "科室": "",
    "检查名称": "",
    "检查部位": "",
    "指标": [],
    \textcolor{blue}{"检查描述":} "静脉麻醉，持续吸氧、心电监护下肠镜检查：患者左侧卧位，行肛门指诊，进指5cm，未触及明确肿物，退指指套无染血。肠道准备一般，中等黄色粪水残留肠腔，充分吸引后观察。钩拉法循腔插镜至回肠末段10cm，进镜顺利，所见部分末段回肠黏膜光整。回盲瓣呈唇型，阑尾开口清晰。所见结肠粘膜光滑，血管纹理清晰，无充血、糜烂、溃疡及新生物。直肠粘膜未见异常。可见轻度内痔。",
    "诊断": "",
    "处理建议": "",
    "样本采集时间": "",
    "其他": ""
\}
```
</answer>
& \{
        "姓名": "",
        "性别": "",
        "年龄": "",
        "检查时间": "",
        "科室": "",
        "检查名称": "",
        \textcolor{red}{"检查部位":} "肠道",
        "指标": [],
        "检查描述": "钩拉法循腔插镜至回肠末段10cm，进镜顺利，所见部分末段回肠黏膜光整。回盲瓣呈唇型，阑尾开口清晰。所见结肠粘膜光滑，血管纹理清晰，无充血、糜烂、溃疡及新生物。直肠粘膜未见异常。可见轻度内痔。",
        "诊断": "",
        "处理建议": "",
        "样本采集时间": "",
        \textcolor{red}{"其他":} "肠道准备一般，中等黄色粪水残留肠腔，充分吸引后观察。"
    \} \\
    \hline
    F1 & 1.0  & 0.0 \\
    Precision & 1.0 & 0.0 \\
    Recall & 1.0 & 0.0 \\
    TED Acc& 1.0 & 0.52 \\

  \end{tabularx}
  }
  \caption{\textbf{Model performances of one case.} Blue letters stands for good parts in model's thinking and output contexts, and red letters stands for wrong contents given by model, revealing that thinking progress makes model do VIE tasks better.}
  \label{tab:thinkcase1}
\end{table}

\begin{table}[!h]
  \centering

\begin{center}
  \includegraphics[width=0.6\columnwidth]{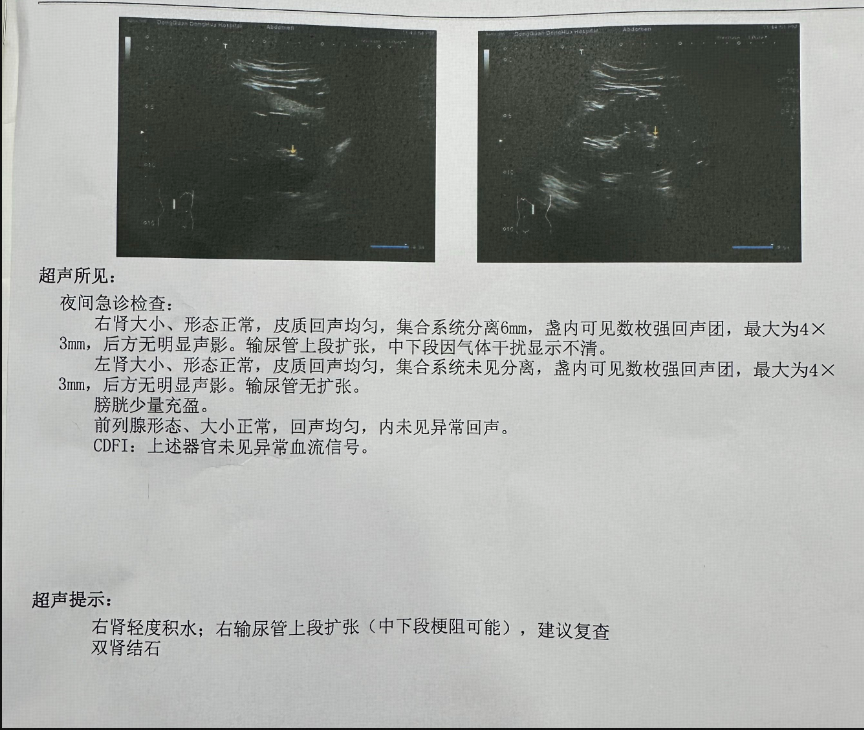}
  
  \vspace{2pt}
  
   
   \end{center}

  \scalebox{0.7}{
  \begin{tabularx}{\textwidth}{X|X|X}
    \hline
    &\textbf{RL-100 (w/ think)} & \textbf{RL-100 (w/o think)}  \\
    \hline 
    Model Output&<think>
从图片内容来看，这是一份超声检查报告。\textcolor{blue}{报告中没有明确提到患者的姓名、性别、年龄、检查时间、科室、检查名称、检查部位、样本采集时间等信息。因此，这些信息无法从图片中提取。}报告中主要描述了超声检查的结果和诊断。报告中提到的检查结果包括右肾轻度积水、右输尿管上段扩张（中下段梗阻可能）和双肾结石。\textcolor{blue}{报告中没有单独列出的检查名称、检查部位、指标、检测方法、结果状态、临床提示、是否为危急值等信息。因此，这些信息也无法从图片中提取。}
</think>
<answer>
\{
    "姓名": "",
    "性别": "",
    "年龄": "",
    "检查时间": "",
    "科室": "",
    "检查名称": "",
    "检查部位": "",
    "指标": [],
    \textcolor{blue}{"检查描述":} "夜间急诊检查：右肾大小、形态正常，皮质回声均匀，集合系统分离6mm，盏内可见数枚强回声团，最大为4×3mm，后方无明显声影。输尿管上段扩张，中下段因气体干扰显示不清。左肾大小、形态正常，皮质回声均匀，集合系统未见分离，盏内可见数枚强回声团，最大为4×3mm，后方无明显声影。输尿管无扩张。膀胱少量充盈。前列腺形态、大小正常，回声均匀，内未见异常回声。CDFI：上述器官未见异常血流信号。",
    \textcolor{blue}{"诊断":} "右肾轻度积水；右输尿管上段扩张（中下段梗阻可能），建议复查双肾结石",
    "处理建议": "",
    "样本采集时间": "",
    "其他": ""
\}
</answer>
& \{
        "姓名": "",
        "性别": "",
        "年龄": "",
        \textcolor{red}{"检查时间":} "夜间急诊检查",
        "科室": "",
        "检查名称": "",
        "检查部位": "",
        "指标": [],
        "检查描述": "右肾大小、形态正常，皮质回声均匀，集合系统分离6mm，盏内可见数枚强回声团，最大为4×3mm，后方无明显声影。输尿管上段扩张，中下段因气体干扰显示不清。左肾大小、形态正常，皮质回声均匀，集合系统未见分离，盏内可见数枚强回声团，最大为4×3mm，后方无明显声影。输尿管无扩张。膀胱少量充盈。前列腺形态、大小正常，回声均匀，内未见异常回声。CDFI：上述器官未见异常血流信号。",
        "诊断": "右肾轻度积水；右输尿管上段扩张（中下段梗阻可能），建议复查双肾结石",
        \textcolor{red}{"处理建议":} "双肾结石",
        "样本采集时间": "",
        \textcolor{red}{"其他":} "检查结果未见异常血流信号"
    \} \\
    \hline
    F1 & 0.4  & 0.0 \\
    Precision & 0.5 & 0.0 \\
    Recall & 0.33 & 0.0 \\
    TED Acc& 0.96 & 0.87 \\

  \end{tabularx}
  }
  \label{tab:thinkcase2}
  \caption{\textbf{Model performances of one case.} Blue letters stands for good parts in model's thinking and output contexts, and red letters stands for wrong contents given by model, revealing that thinking progress makes model do VIE tasks better.}
\end{table}

\end{CJK}
\end{document}